# One-Step or Two-Step Optimization and the Overfitting Phenomenon: A Case Study on Time Series Classification

Muhammad Marwan Muhammad Fuad

*Forskningsparken 3, Institutt for kjemi, NorStruct*
*The University of Tromsø - The Arctic University of Norway, NO-9037 Tromsø, Norway*
*mfu008@post.uit.no*

Abstract: For the last few decades, optimization has been developing at a fast rate. Bio-inspired optimization algorithms are metaheuristics inspired by nature. These algorithms have been applied to solve different problems in engineering, economics, and other domains. Bio-inspired algorithms have also been applied in different branches of information technology such as networking and software engineering. Time series data mining is a field of information technology that has its share of these applications too. In previous works we showed how bio-inspired algorithms such as the genetic algorithms and differential evolution can be used to find the locations of the breakpoints used in the symbolic aggregate approximation of time series representation, and in another work we showed how we can utilize the particle swarm optimization, one of the famous bio-inspired algorithms, to set weights to the different segments in the symbolic aggregate approximation representation. In this paper we present, in two different approaches, a new meta optimization process that produces optimal locations of the breakpoints in addition to optimal weights of the segments. The experiments of time series classification task that we conducted show an interesting example of how the overfitting phenomenon, a frequently encountered problem in data mining which happens when the model overfits the training set, can interfere in the optimization process and hide the superior performance of an optimization algorithm.

## 1 INTRODUCTION

For the last few decades, optimization has been developing at a fast rate. Novel algorithms appear and new applications emerge in different fields of engineering, economics, and science. The purpose of an optimization process is to find the best-suited solution of a problem subject to given constraints. These constraints can be in the boundaries of the parameters controlling the optimization problem, or in the function to be optimized. This can be expressed mathematically as follows: Let $\vec{X} = [x_1, x_2, ..., x_{nbp}]$ be the candidate solution to the problem for which we are searching an optimal solution. Given a function $f : U \subseteq \mathbf{R}^{nbp} \rightarrow \mathbf{R}$ (*nbp* is the number of parameters), find the solution $\vec{X}^* = [x_1^*, x_2^*, ..., x_{nbp}^*]$ which satisfies:

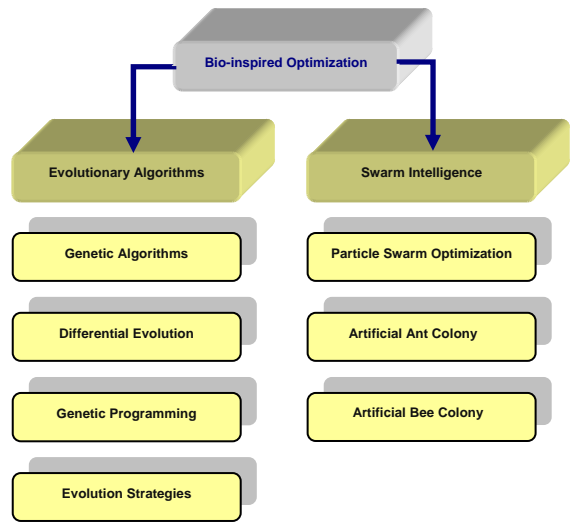

Figure 1: Some of the main bio-inspired metaheuristics.

$f\left(\vec{X}^*\right) \leq f\left(\vec{X}\right), \forall \vec{X} \in U$. The function *f* is called the *fitness function*, or the *objective function* (These two concepts are not really identical, but in this paper we will use them interchangeably).

*Metaheuristics* are probabilistic optimization algorithms which are applicable to a large variety of optimization problems. Metaheuristics can be divided into *single solution –based* metaheuristics and *population-based* metaheuristics. Many of these metaheuristics are inspired by natural processes, natural phenomena, or by the collective intelligence of natural agents, hence the term *bio-inspired,* also called *nature-inspired*, optimization algorithms. Figure 1 shows the main bio-inspired metaheuristics.

One of the main bio-inspired optimization families is *Evolutionary Algorithms* (EA). EA are population-based metaheuristics that use the mechanisms of Darwinian evolution. The *Genetic Algorithm* (GA) is the main member of EA. GA is an optimization and search technique based on the principles of genetics and natural selection (Haupt and Haupt, 2004). GA has the following elements: a population of individuals, selection according to fitness, crossover to produce new offspring, and random mutation of new offspring (Mitchell, 1996).

*Differential Evolution* (DE) is an optimization method which is mainly effective to solve continuous problems. DE is also based on the principles of genetics and natural selection.

DE has the same elements as a standard evolutionary algorithm; i.e. a population of individuals, selection according to fitness, crossover, and random mutation. DE starts with a collection of randomly chosen individuals constituting a population whose size is *popsize*. Each of these solutions is a vector of *nbp* dimensions and it represents a possible solution to the problem at hand. The fitness function of each individual is evaluated. The next step is optimization. In this step for each individual, which we call the *target vector* $\vec{T}_i$, of the population three mutually distinct individuals $\vec{V}_{r1}, \vec{V}_{r2}, \vec{V}_{r3}$, and different from $\vec{T}_i$, are chosen at random from the population. The *donor vector* $\vec{D}$ is formed as a weighted difference of two of $\vec{V}_{r1}, \vec{V}_{r2}, \vec{V}_{r3}$ added to the third; i.e. $\vec{D} = \vec{V}_{r1} + F(\vec{V}_{r2} - \vec{V}_{r3})$. *F* is called the *mutation factor* or the *differentiation constant* and it is one of the *control parameters* of DE.

The *trial vector* $\vec{R}$ is formed from elements of the target vector $\vec{T}_i$ and elements of the donor vector $\vec{D}$. In the following we present the crossover scheme presented in (Feoktistov, 2006) which we adopt in this paper; an integer *Rnd* is randomly chosen among the dimensions $[1, nbp]$. Then the trial vector $\vec{R}$ is formed as follows:

$$t_i = \begin{cases} t_{i,r1} + F(t_{i,r2} - t_{i,r3}) \\ \quad if \ (rand_{i,j}[0,1[ < C_r) \vee (Rnd = i) \\ t_{i,j} \\ \quad otherwise \end{cases} \quad (1)$$

where $i = 1,...,nbp$. $C_r$ is the *crossover constant*, which is another control parameter.

The next step of DE is selection. This step decides which of the trial vector and the target vector will survive in the next generation and which will die out. The selection is based on which of the two vectors; trial and target, yields a better value of the fitness function. Crossover and selection repeat for a certain number of generations.   □

Data mining is a field of computer science which handles several tasks such as classification, clustering, anomaly detection, and others. Processing these tasks usually requires extensive computing. As with other fields of computer science, different papers have proposed applying bio-inspired optimization to data mining tasks (Muhammad Fuad, 2012a, 2012b, 2012c, 2012d).

Data mining models may suffer from what is called *overfitting*. A classification algorithm is said to *overfit* to the training data if it generates a representation of the data that depends too much on irrelevant features of the training instances, with the result that it performs well on the training data but relatively poorly on unseen instances (Bramer, 2007). In overfitting the algorithm memorizes the training set at the expense of generalizability to the validation set (Larose, 2005).

We will show in this paper a generalization of the optimization methods that we presented in (Muhammad Fuad, 2012b, 2012c, 2012d) and how the overfitting phenomenon can interfere to alter the outcome of the optimization process. In Section 2 we present the problem and our previous work on it and the new generalization, in two different approaches, of this previous work. We compare these two approaches in Section 3 and we discuss the results of the experiments in Section 4. We conclude the paper in Section 5.

## 2 ONE-STEP OR TWO-STEP OPTIMIZATION OF THE SYMBOLIC AGGREGATE APPROXIMATION

A *time series S* is an ordered collection:

$$S = \{(t_1, v_1), (t_2, v_2), ..., (t_n, v_n)\} \quad (2)$$

where $t_1 < t_2 < ... < t_n$, and where $v_i$ are the values of the observed phenomenon at time points $t_i$.

Time series data mining handles several tasks such as classification, clustering, similarity search, motif discovery, anomaly detection, and others.

Time series are high-dimensional data so they are usually processed by using representation methods that are used to extract features from these data and project them on lower-dimensional spaces.

The *Symbolic Aggregate approXimation* method (SAX) (Lin *et. al.*, 2003) is one of the most important representation methods of time series. The main advantage of SAX is that the similarity measure it uses, called MINDIST, uses statistical lookup tables, which makes it easy to compute with an overall complexity of $O(N)$. SAX is based on the assumption that normalized time series have Gaussian distribution, so by determining the breakpoints that correspond to a particular alphabet size, one can obtain equal-sized areas under the Gaussian curve. SAX is applied as follows:

1- The time series are normalized.
2- The dimensionality of the time series is reduced using PAA (Keogh *et. al.*, 2000), (Yi and Faloutsos, 2000).
3- The PAA representation of the time series is discretized by determining the number and location of the breakpoints (The number of the breakpoints is chosen by the user). Their locations are determined, as mentioned above, using Gaussian lookup tables. The interval between two successive breakpoints is assigned to a symbol of the alphabet, and each segment of PAA that lies within that interval is discretized by that symbol.

The last step of SAX is using the following similarity measure:

$$MINDIST(\hat{S}, \hat{R}) \equiv \sqrt{\frac{n}{N}} \sqrt{\sum_{i=1}^{N} (dist(\hat{s}_i, \hat{r}_i))^2} \quad (3)$$

Where $n$ is the length of the original time series, $N$ is the length of the strings (the number of the segments), $\hat{S}$ and $\hat{R}$ are the symbolic representations of the two time series $S$ and $R$, respectively, and where the function $dist(\ )$ is implemented by using the appropriate lookup table. It is proven that the similarity measure in (3) produces no false dismissals.

In (Muhammad Fuad, 2012b), (Muhammad Fuad, 2012c), we showed that this assumption of Gaussianity oversimplifies the problem and can result in very large errors in time series mining tasks, and we presented alternatives to SAX; one which is based on the genetic algorithms (GASAX), and another which is based on the differential evolution (DESAX), to determine the locations of the breakpoints. We showed that these two optimized alternatives clearly outperform the original SAX.

In (Muhammad Fuad, 2012d), we used *Particle Swarm Optimization* (PSO), well-known metaheuristics, to optimize SAX using a different approach; to propose a new minimum distance WMD that can better recover the information loss resulting from SAX. The method we introduced, (PSOWSAX), gives different weights to different segments of the times series in the lower-dimensional space. These weights are proportional to the information contents of the different segments. In (PSOWSAX) PSO is utilized to set these weights.

A logical extension, which we present in this work, of the above optimized alternatives to SAX is one meta optimization process that can find optimal locations of the breakpoints and optimal weights of the segments corresponding to these breakpoints. Such an optimization process can actually be handled in two different ways; the first, which we call *Two-Step OSAX* constitutes of two steps, as the name implies. In the first step the optimization searches for the optimal locations of the breakpoints assuming that all segments have the same weight $w_i = 1$, and then in the next step another optimization process sets optimal weights based on the optimal locations of the breakpoints, which are the outcome of step one. Since the weights are related to the locations of the breakpoints, it is logical that step one should start by optimizing the locations of the breakpoints. The other way, which we call *One-Step OSAX* uses one step to find the optimal locations of the breakpoints and the weights corresponding to those breakpoints.

We tended to think before conducting the experiments that *Two-Step OSAX* will give better results than *One-Step OSAX* since it seemed more intuitive. We will see in the experimental section if this intuition is founded.

## 3 EXPERIMENTS

We conducted the same experiments that we used to validate (GESAX), (DESAX), and (PSOWSAX), which are *1*-NN classification task experiments. The goal of classification, one of the main tasks of data mining, is to assign an unknown object to one out of a given number of classes, or categories (Last, Kandel, and Bunke , 2004). In *k*-NN each time series is assigned a class label. Then a "leave one out" prediction mechanism is applied to each time series in turn; i.e. the class label of the chosen time series is predicted to be the class label of its nearest neighbor, defined based on the tested distance function. If the prediction is correct, then it is a hit; otherwise, it is a miss.

The classification error rate is defined as the ratio of the number of misses to the total number of time series (Chen and Ng, 2004). *1*-NN is a special case of *k*-NN, where *k=1*.

We conducted experiments using datasets of different sizes and dimensions available at UCR (Keogh *et. al.*, 2011).

The values of the alphabet size we tested in the experiments were 3, 10, and 20. We chose these values because the first version of SAX used alphabet size that varied between 3 and 10. Then in a later version the alphabet size varied between 3 and 20. So these values are benchmark values.

As for the control parameters of DE that we used, the population size *popsize* was 12. The differentiation constant $F_r$ was set to 0.9, and the crossover constant $C_r$ was set to 0.5. The dimension of the problem *nbp* is *breakpoints + weights* in *One-Step OSAX*, and it is *breakpoints* and *weights*, in the first step of *Two-Step OSAX* and the second step of *Two-Step OSAX*, respectively. As for the number of generations *NrGen*, it is 100 for *One-Step OSAX* and 50 for each step of *Two-Step OSAX*. Since the fitness function is the same (the classification error) this configuration guarantees that the complexities of *One-Step OSAX* and that of *Two-Step OSAX* are almost the same.

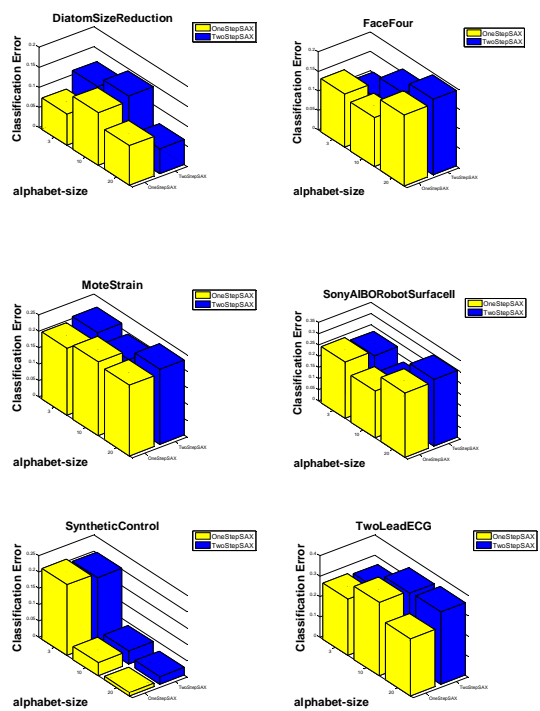

Figure 2: Comparison of the classification errors between *One-Step OSAX* and *Two-Step OSAX* on different testing datasets.

The training phase for *Two-Step OSAX* is as follows; for each value of the alphabet size, we run the optimization process for 50 generations to get the optimal locations of the breakpoints. This corresponds to the first step of *Two-Step OSAX*, then we use these optimal locations of the breakpoints to run the optimization process for 50 generations to obtain the optimal weights. Later we use these optimal weights and locations of the breakpoints on the testing datasets to obtain the minimal classification error (which we show in Figure 2 for some of the different datasets that we tested in our experiments). As for the experimental protocol for *One-Step OSAX*, it is the same as that for *Two-Step OSAX*, except that we optimize in one step for the locations of the breakpoints and for the weights, also *NrGen* in the case is equal to 100.

The results shown in Figure 2 do not show, in general, a difference in performance between *One-Step OSAX* and *Two-Step OSAX*. We will discuss these results more thoroughly in the next section.

# 4 DISCUSSION

At first glance the results we obtained in Section 3 suggest that there is no real difference between the performance of *One-Step OSAX* and that of *Two-Step OSAX*. We have to admit that these results were a bit surprising to us as we expected, before conducting the experiments, that *Two-Step OSAX* would give better results than those obtained by applying *One-Step OSAX* for reasons we mentioned in Section 2. However, a deeper examination of the results may yet reveal more surprising results.

If we remember from the experimental protocol illustrated in the previous section, the two optimization processes: *One-Step OSAX* and *Two-Step OSAX* are applied to the training datasets to find the optimal locations of the breakpoints and the optimal weights, and then these optimal values are applied to the testing datasets, so the real comparison between the efficiency of *One-Step OSAX* and *Two-Step OSAX*, as optimization processes, should in fact compare their performances on the training datasets. Table 1 shows the classification error; the fitness function in our optimization problem, obtained by applying *One-Step OSAX* and *Two-Step OSAX* to the datasets presented in Figure 2. The results presented in Table 1 show that *One-Step OSAX* clearly outperforms *Two-Step OSAX*. These results raise two questions; the first is why these results did not appear in Figure 2, which represents the final outcome of the whole process, and the second question is why *One-Step OSAX* outperformed *Two-Step OSAX*. As for the first and the most interesting question which is why the superior performance of *One-Step OSAX* did not show in the final results presented in Figure 2, the reason for this is that the optimal solutions obtained by *One-Step OSAX* on the training datasets overfitted the data; they were "too good" that when they were used on the testing datasets the performance dropped. This was not the case with *Two-Step OSAX*. This interesting finding may establish a new strategy to applying optimization in data mining that "less optimal" solutions obtained by applying the optimization algorithm to the training datasets may finally turn out to be better than "more optimal" solutions which may overfit the training data.

As for the other question of why *One-Step OSAX* outperformed *Two-Step OSAX*, we believe the reason for this could be that finally the locations of the breakpoints play a more important role in the optimization process than the weights, so in *One-Step OSAX* the optimization process kept improving the locations of the breakpoints until the final stages of the optimization process while *Two-Step OSAX* interrupted this optimization after the first step.

Table 1: Comparison of the classification errors between *One-Step OSAX* and *Two-Step OSAX* on the training datasets presented in Figure 2.

| Dataset | Method | Classification Error | | |
| --- | --- | --- | --- | --- |
| | | $\alpha^*=3$ | $\alpha=10$ | $\alpha=20$ |
| DiatomSizeReduction | One-Step OSAX | 0.062 | 0.062 | 0.062 |
| | Two-Step OSAX | 0.687 | 0.125 | 0.062 |
| FaceFour | One-Step OSAX | 0.042 | 0.042 | 0.167 |
| | Two-Step OSAX | 0.292 | 0.167 | 0.167 |
| MoteStrain | One-Step OSAX | 0.150 | 0.050 | 0.050 |
| | Two-Step OSAX | 0.700 | 0.150 | 0.100 |
| SonyAIBORobotSurfaceII | One-Step OSAX | 0.148 | 0.111 | 0.074 |
| | Two-Step OSAX | 0.444 | 0.185 | 0.148 |
| synthetic_control | One-Step OSAX | 0.170 | 0.003 | 0.000 |
| | Two-Step OSAX | 0.833 | 0.020 | 0.003 |
| TwoLeadECG | One-Step OSAX | 0.130 | 0.087 | 0.043 |
| | Two-Step OSAX | 0.652 | 0.260 | 0.087 |

$\alpha^*$ is the alphabet size

## 5 CONCLUSIONS

In this paper we presented two approaches of an optimization algorithm that handles a certain problem of time series classification. These time series are represented using a symbolic representation method, SAX, of time series data. Our optimized versions produce optimal versions, in terms of classification task accuracy, of the original SAX in that they spot optimal locations of the breakpoints and optimal weights compared with the original SAX. The first of the two approaches we presented, *One-Step OSAX*, consists of one step to perform the optimization process, while the second one, *Two-Step OSAX*, applies the optimization process in two steps. The experiments we conducted show that *One-Step OSAX* outperforms *Two-Step OSAX*. The interesting finding of our experiments is that the superior performance of *One-Step OSAX* was actually "hidden" because of the overfitting phenomenon.

We believe the results we obtained shed new light on the application of optimization algorithms, mainly stochastic and bio-inspired ones, in data mining, where the risk of overfitting is frequently present. We think that an optimization algorithm, applied to the training data, should avoid searching for very close-to-optimal solutions, which highly fit the training data and thus may not give optimal solutions when deployed to testing data. A possible strategy that we suggest is to terminate the optimization process, on the training sets, prematurely. However, the output of our experiments applies to times series only, and to the classification task in particular, so we do not have enough evidence that our remarks are generalizable.